\documentclass[11pt]{article}

\usepackage[final]{acl}

\usepackage{times}
\usepackage{latexsym}
\usepackage{xcolor}
\usepackage{inconsolata}

\usepackage{graphicx}

\usepackage{longtable}
\usepackage{color,soul}
\usepackage[T1, T2A]{fontenc}
\usepackage[utf8]{inputenc} 
\usepackage{tempora}
\usepackage{hyperref}       
\usepackage{url}            
\usepackage{booktabs}       
\usepackage{amsfonts}       
\usepackage{nicefrac}       
\usepackage{microtype}      
\usepackage{xcolor}         
\usepackage[utf8]{inputenc}
\usepackage[T1]{fontenc}
\usepackage{adjustbox}
\usepackage{dirtytalk}

\title{The PLLuM Instruction Corpus}

\author{
 \textbf{Piotr Pęzik\textsuperscript{1}},
 \textbf{Filip Żarnecki\textsuperscript{1}},
 \textbf{Konrad Kaczyński\textsuperscript{1}},
 \textbf{Anna Cichosz\textsuperscript{1}},
 \textbf{Zuzanna Deckert\textsuperscript{1}},
 \\
 \textbf{Monika Garnys\textsuperscript{1}},
 \textbf{Izabela Grabarczyk\textsuperscript{1}},
 \textbf{Wojciech Janowski\textsuperscript{1}},
 \textbf{Sylwia Karasińska\textsuperscript{1}}, 
\\
 \textbf{Aleksandra Kujawiak\textsuperscript{1}},
 \textbf{Piotr Misztela\textsuperscript{1}}, 
 \textbf{Maria Szymańska\textsuperscript{1}},
 \textbf{Karolina Walkusz\textsuperscript{1}}, 
 \textbf{Igor Siek\textsuperscript{1}}, 
\\
 \textbf{Maciej Chrabąszcz\textsuperscript{2}},
 \textbf{Anna Kołos\textsuperscript{2}},
 \textbf{Agnieszka Karlińska\textsuperscript{2}},
 \textbf{Karolina Seweryn\textsuperscript{2}},
\\ 
 \textbf{Aleksandra Krasnodębska\textsuperscript{2}}, 
 \textbf{Paula Betscher\textsuperscript{2}},
 \textbf{Zofia Cieślińska\textsuperscript{2}},
 \textbf{Katarzyna Kowol\textsuperscript{2}},
\\
  \textbf{Artur Wilczek\textsuperscript{2}},
 \textbf{Maciej Trzciński\textsuperscript{2}},
 \textbf{Katarzyna Dziewulska\textsuperscript{2}},
 \textbf{Roman Roszko\textsuperscript{3}},
 \\
 \textbf{Tomasz Bernaś\textsuperscript{3}}, 
 \textbf{Jurgita Vaičenonienė\textsuperscript{3}},
 \textbf{Danuta Roszko\textsuperscript{3}},
 \textbf{Paweł Levchuk\textsuperscript{3}},
 \textbf{Paweł Kowalski\textsuperscript{3}},
\\
 \textbf{Irena Prawdzic-Jankowska\textsuperscript{3}},
 \textbf{Marek Kozłowski\textsuperscript{4}},
 \textbf{Sławomir Dadas\textsuperscript{4}},
 \textbf{Rafał Poświata\textsuperscript{4}},
\\
 \textbf{Alina Wróblewska\textsuperscript{5}},
 \textbf{Katarzyna Krasnowska-Kieraś\textsuperscript{5}},
\textbf{Maciej Ogrodniczuk\textsuperscript{5}},
\textbf{Michał Rudolf\textsuperscript{5}},
\\
\textbf{Piotr Rybak\textsuperscript{5}},
\textbf{Karolina Saputa\textsuperscript{5}},
\textbf{Joanna Wołoszyn\textsuperscript{5}},
\textbf{Marcin Oleksy\textsuperscript{6}},
\textbf{Bartłomiej Koptyra\textsuperscript{6}},
\\
\textbf{Teddy Ferdinan\textsuperscript{6}},  
\textbf{Stanisław Woźniak\textsuperscript{6}}, 
\textbf{Maciej Piasecki\textsuperscript{6}},  
\textbf{Paweł Walkowiak\textsuperscript{6}},
\\
\textbf{Konrad Wojtasik\textsuperscript{6}},
\textbf{Arkadiusz Janz\textsuperscript{6}},
\textbf{Przemysław Kazienko\textsuperscript{6}}, 
\textbf{Julia Moska\textsuperscript{6}}, 
\textbf{Jan Kocoń\textsuperscript{6}}
\\
\\
 \textsuperscript{1} University of Lodz\\
 \textsuperscript{2} NASK National Research Institute\\
 \textsuperscript{3} Institute of Slavic Studies PAS\\
 \textsuperscript{4} National Information Processing Institute\\
 \textsuperscript{5} Institute of Computer Science PAS\\
 \textsuperscript{6} Wroclaw Tech
\\
 \small{
   \textbf{Correspondence:} \href{mailto:piotr.pezik@uni.lodz.pl}{piotr.pezik@uni.lodz.pl}
 }
}

\begin{document}
\maketitle
\begin{abstract}
This paper describes the instruction dataset used to fine-tune a set of transformer-based large language models (LLMs) developed in the PLLuM (Polish Large Language Model) project. We present a functional typology of the organic, converted, and synthetic instructions used in PLLuM and share some observations about the implications of using human-authored versus synthetic instruction datasets in the linguistic adaptation of base LLMs. Additionally, we release the first representative subset of the PLLuM instruction corpus (PLLuMIC), which we believe to be useful in guiding and planning the development of similar datasets for other LLMs.
\end{abstract}

\section {Introduction}

The Polish Large Language Model (PLLuM) was a project funded by the Polish Ministry of Digital Affairs in 2024\footnote{See \url{http://pllum.org.pl}.}. Its main delivery was a `family' of language-adapted, fine-tuned, and aligned large language models (LLMs) ranging in size from 8 to 70 billion parameters as summarized in Table \ref{tab:pllum_models}. One of the central tasks of the project was the design of an original corpus of instructions that could be used to develop the basic interactive capabilities of the target models. Several challenges that became apparent in creating such a~dataset formed the core motivation for this study. 

First, the datasets used to fine-tune both proprietary and open-weight LLMs are usually withheld by their developers. This, in turn, makes the replication of LLM capabilities difficult. Second, the composition of stand-alone instruction datasets (i.e., datasets released independently of any specific LLM) is usually poorly documented. Such resources are typically constructed opportunistically, often as a conflation of other datasets, and even when they follow a predefined typology, the accompanying documentation is often too sparse to serve as a reliable foundation for designing similar datasets for LLM development.

Furthermore, while the role of instruction datasets in LLM development is generally acknowledged, there is relatively little published research on how different types of instructions impact the capabilities of original, published models. Another clear research gap is related to the growing trend of large-scale instruction distillation from so-called \emph{strong LLMs}. While this approach offers a convenient shortcut to creating instruction datasets, it comes with unobvious limitations, especially in the context of linguistic and cultural adaptation of LLMs, which was central to the PLLuM project. Paradoxically, there is also the converse issue of human authors of instructions having to learn the style of LLM responses, which has recently emerged as a new register of language\footnote{We use the term register in the sense of a functional genre or variety of language \cite{conrad_register_2023}.} as a result of human interactions with popular LLMs. Finally, deriving instructions from annotated corpora and structured knowledge sources (i.e., \emph{automatic instructions}), while promising in many respects, introduces the risk of distorting the overall balance between organic, automatic, and synthetic instructions (see our definitions below) in the dataset and consequently also the behaviour of the resulting model.\footnote{Automatically converted instructions tend to be very repetitive. In large quantities, they may affect the conversational fluency of a fine-tuned LLM.}

This paper presents the instruction datasets used to fine-tune the PLLuM models. We describe our functional typology of the organic, automatic, and synthetic instructions in the context of the LLM research issues signalled above. We also release a representative subset of the PLLuM instruction corpus (PLLuMIC) to provide potential guidance and inspiration for developing similar datasets for other LLMs.

\begin{table*}
  \centering
  \begin{tabular}{ll}
    \toprule
    \textbf{Model name} & \textbf{Base model} \\
    \cmidrule(r){1-2}
    PLLuM-12B-nc-chat & Mistral-Nemo-Base-2407 \\
    PLLuM-8x7B-nc-chat & Mixtral-8x7B-v0.1 \\
    Llama-PLLuM-8B-chat & Llama-3.1-8B \\
    Llama-PLLuM-70B-chat &Llama-3.1-70B \\
    \bottomrule
  \end{tabular}
  \caption{A subset of the models adapted, fine-tuned, and aligned in PLLuM.}
  \label{tab:pllum_models}
\end{table*}

\section{LLM Fine-tuning}

In the context of Large Language Model (LLM) development, the term 
\textit{instructions} refers to single- or multi-turn question-and-answer pairs, i.e. $(Q_i, A_i)$, which exemplify the format, style, and functional content of interactions between the model and its users:
\begin{equation}
  \label{eq:example}
    I = \{(Q_1, A_1), (Q_2, A_2), \dots, (Q_n, A_n)\} 
\end{equation}

Instructions can be categorized with respect to their origin as:

\begin{itemize}
  \item \textit{Organic}, i.e.\ authored by humans, including experts and trained annotators. Manual instructions can also be crowd-sourced or collected from human prompts in interactions with existing LLMs.
  \item \textit{Converted}, i.e., derived from annotated corpora, knowledge sources, etc.
  \item \textit{Synthetic}, i.e.\ distilled more or less directly from existing LLMs through manual or automated prompting techniques.
\end{itemize}

Hybrid scenarios for acquiring instructions are also possible in that synthetic instructions can be verified and corrected manually, organic and automatically converted instructions can be enhanced by LLMs, etc. As we explain below, each of these three basic sources of acquiring instructions has its advantages and limitations.

The fine-tuning of pre-trained models on instruction datasets remains a crucial step in the de\-ve\-lop\-ment of LLMs based on the transformer architecture. Although base models can perform certain tasks and interpolate between knowledge items attested in pre-training, it is clear that balanced, high-quality datasets of instructions are indispensable resources in text-to-text LLM development workflows \cite{longpre2023flan}. 

\section{Availability and Transparency of Instruction Datasets}

Although the basic steps of developing LLMs, such as fine-tuning on instructions, are widely researched, there is relatively little practical information about the composition of datasets used in real-world model-building projects. For various legal and business-related reasons, high-quality text corpora, instructions, and preferences are often withheld or inadequately documented by vendors and publishers of closed and open LLMs. We provide an overview of the transparency of instruction datasets used to develop a number open-weight models in Appendix \ref{sec:appendix_transparency}\footnote{A useful distinction is made between open-source and open-weight models, where the latter are usually provided without key data resources.} In short, the vast majority of such models are provided without the instructions used to fine-tune them and with very little if any documentation about such resources. 

On the other hand, open instruction datasets (often developed independently of any particular LLM), tend to be largely opportunistic \footnote{Appendix \ref{sec:appendix_transparency} contains a summary of open instruction datasets availability and documentation.}. The definition and compilation of balanced instruction corpora remains a major methodological challenge for any team developing an original instruction fine-tuned LLM. Even in projects which utilize large-scale distillation of skills and knowledge from existing models, a general functional typology of human-LLM interactions is required to design the corpus of instructions.

\section{The Composition of PLLuMIC}
In the following section, we introduce the structure of the PLLuM Instruction Corpus (henceforth PLLuMIC). Although by design, the bulk of the corpus consists of (1) hand-crafted, high-quality organic instructions curated by a team of trained annotators, we also explored the value of (2) instructions distilled from existing LLMs and (3) converted from annotated corpora database and text repositories.

\begin{table}
\renewcommand\tabcolsep{17pt}

  \centering
  \begin{tabular}{lc}
    \toprule
    \textbf{Category} & \textbf{Proportion} \\
    \midrule
    \verb|Knowledge (QA)|               & 43\%          \\
    \verb|Generation|                   & 25\%          \\
    \verb|Extraction|                   & 6\%           \\
    \verb|Programming|                  & 6\%           \\
    \verb|Conversational|               & 4\%           \\
    \verb|NLP|                          & 3\%           \\
    \verb|Adversarial|                  & 3\%           \\
    \verb|Visualization|                & 3\%           \\
    \verb|Data manipulation|            & 3\%           \\
    \verb|Chain of Thought|             & 2\%           \\
    \verb|Translation|                  & 1\%           \\
    \verb|Identity|                     & 1\%           \\
    \bottomrule
  \end{tabular}
  \caption{High-level PLLuMIC composition with their respective approximate representation in the full organic component of the corpus.}
  \label{tab:table_high_level_composition}
\end{table}

\subsection{Organic Instructions}\label{subsec:organic_ins}
Instructions annotated by professional human annotators hired for the project formed the primary component of PLLuMIC. We refer to such instructions as \emph{organic} to distinguish them from synthetic and automatic or `converted' instructions. They were either written from scratch by single or multiple annotators to fill the above-mentioned categories or adapted from open datasets. Adaptation of open-source datasets (e.g. CREAK \cite{onoe2021creak} (3591 samples), ECQA \cite{aggarwaletal2021ecqa} (1033 samples), QED \cite{lamm2020qed} (1855 samples)) was an effective initial strategy, but this approach showed significant limitations with time. Many of the adapted samples contained low-quality, simplistic, or erroneous instructions, but with some effort invested in their corrections, they proved to be useful for both fine-tuning and evaluation purposes. 

The annotation process was subject to rigorous quality control, described in more detail in Appendix \ref{subsec:quality_control}.

\paragraph{Prompt-response Instructions} We started the core manual annotation phase with a small ad-hoc typology covering mostly simple prompt-response interactions such as factual knowledge and commonsense reasoning question-answering and several generative subtypes, i.e. short text composition prompts with relatively long expected output. Some extractive tasks, such as summarization and keyphrase identification, were also considered in this initial phase. The resulting high-level composition of PLLuMIC is outlined in Table \ref{tab:table_high_level_composition}. A more detailed account of the PLLuMIC typology is given in Appendix \ref{sec:appendix_typology}.

\paragraph{Dialogue instructions}
One of the stages in the development of PLLuMIC was the shift from simple prompt-response instructions to multi-turn dialogues. Although prompt-response turns are prototypical instructions, they fail to capture more sophisticated conversational scenarios such as role-playing, context-sensitivity, and multi-turn prompting, whereby several stages of interaction are required to specify and solve a task at hand. As the dataset size became sufficient to fine-tune early versions of our LLMs, instructions and multi-turn dialogues were also gathered through human-model interactions. The responses generated by intermediate fine-tuned models were carefully validated and refined before being included in the instruction dataset. We created a subset of over 3,500 dialogues with an average of approximately 12 turns per conversation. 

A subset of our typology also features instructions in other languages, mostly Ukrainian, Lithuanian, Russian and Belarussian.

\subsection{Synthetic Instructions}
\label{sec:comp_synthetic}
To extend the range of tasks and topical domains covered by human annotators, we generated an experimental subset of high-quality instructions using selected LLMs with limited human supervision. To this end, we gradually devised a map of topical domains (Appendix \ref{sec:appendix_synth_knowledge_distilled} \& \ref{sec:appendix_synth_context_injected}), and depending on the type of skill or knowledge, we used different multi-step generation-pipelines involving minimal human supervision and several locally-hosted LLMs. The main two types of synthetic instructions included in PLLuMIC were focused on \textit{Knowledge distillation}, \textit{RAG} and \textit{Context-injected NLP} tasks.

\subsubsection{Knowledge Distillation}
Starting with a manually constructed list of topics and subtopics; for each topic, human annotators compiled a series of hand-written subject prompts that were subsequently injected into a meta-prompt generating a question; then LLM-generated questions were fed into a meta-prompt to generate the answer. Meta-prompts at each pipeline step contained detailed specifications of the desired content, style, and format. All prompt-answer pairs in this phase were generated and validated with the permissively licensed Mixtral8x22b-instruct model.

\subsubsection{RAG Instructions}
\label{rag} 

To optimize PLLuM models for Retrieval Augmented Generation (RAG), especially in the domain of public administration, we compiled a comprehensive set of instructions and preferences from documents available on Polish government websites in the \textit{gov.pl} domain. These included mainly administrative guides, and structured informational pamphlets covering a range of issues such as applying for identity documents, business activity, taxes, residence registration, and others. We prepared three sets of questions: (1) \emph{regular questions} that are likely to be answered by information contained in the indexed documents, (2) \emph{adversarial questions} intended to trick the model into providing unacceptable answers and (3) \emph{unrelated questions}, which were completely unrelated to the topic of the documents and therefore should be ignored. The regular and adversarial questions were generated by a strong LLM for each fragment of the document while unrelated questions were sampled from various QA datasets and reviewed by annotators. Afterwards, for each fragment, the top 5 documents were retrieved via a pipeline composed of the \emph{bge-m3} retriever \footnote{{\label{bge_note}}https://github.com/FlagOpen/FlagEmbedding/tree/master} and the \emph{bge-reranker-v2-m3} reranker \footnotemark[\value{footnote}] \cite{bge_m3}. For each set of questions and retrieved documents, we generated an answer using Llama-3.3-70B (serving as the strong LLM) and treated it as a reference answer. We also generated preferred answers to be used in the model alignment phase, using a weaker model Llama-3.1-8B. To avoid overfitting our generic models, we limited the set of RAG instructions to 5,000 in the SFT phase and 5,000 preferences in the alignment phase. The final training set contained 80\% regular questions, 14\% adversarial questions, and 6\% unrelated questions.
\subsubsection{Context-injected NLP}
Text samples extracted from open-source collections were injected into system prompts containing detailed specifications of NLP tasks, such as named entity recognition, classification, semantic similarity, translation, etc. Special effort was invested in defining the desired structured output formats such as JSON, CSV, XML, etc. Pairs of system prompts and LLM-generated answers were subsequently validated for compliance with the constraints defined inside the system prompt.

\subsubsection{Limitations of Mass-distillation}
Despite the current trend to use large-scale distillation techniques in both LLM training and inference, we attempted to control and mitigate certain synthetic data limitations throughout the development of PLLuMIC. First, many LLMs are governed by licenses restricting data generation for derivative model development. These legal constraints and a lack of expertise in constructing original instruction and alignment datasets may lead to long-term over-dependence on existing LLMs. Second, a distillation of aligned models can propagate biases and pre-existing preferences, potentially compromising our model balancing and neutrality definitions. Furthermore, poorly controlled recursive distillation may lead to model degradation or even collapse \cite{shumailov2024ai}. Finally, as discussed in Section \ref{subs:lang_adapt}, we observe significant negative transfer effects in language-adapted models while transfer learning and task interpolation are fundamental properties of generative language models.

\subsection{Converted Instructions}

Prompt-response pairs can also be automatically created from annotated corpora (e.g., treebanks, named-entity datasets, etc.) and other resources, including machine-readable dictionaries and ontologies. Members of the PLLuM consortium used their experience in developing various types of NLP datasets to convert instances of these datasets into instructions. Responses were often extracted from annotation layers using handwritten question-and-answer templates. For some datasets,\footnote{The text classification tasks from Polish Summaries Corpus, DYK, PolEmo2, Polish CBD, Polish Paraphrase Corpus, CDSC-E, 8tags, and NKJP-NER. More details can be found in Table \ref{tab:converted_taks_examples}.} several prompt formats were prepared. When mapping an example into a single-turn instruction, a prompt format was randomly selected. 

Notably, we only took train splits of these datasets for training, while validation splits were optionally used for internal evaluation.

Table \ref{tab:converted_taks_examples} provides examples of specific subsets of converted instructions. Although this approach allows for the efficient large-scale production of instructions, the resulting data is often highly repetitive, reducing the fine-tuned model's conversational versatility. To mitigate this, we imposed strict limits on the number of converted instructions obtained from each resource.\footnote{By default, only a maximum of 1,000 instructions were converted from a single resource.}

\section{Language Adaptation Experiments}\label{sec:ft_exp}

\begin{table*}[t]

  \centering
  \begin{tabular}{lccc}
    \toprule
    \textbf{Type} & \textbf{Quantity Train}  & \textbf{Proportion Train}  & \textbf{Quantity Total}\\
    \midrule
    Organic     & 38,106  & 49.12\%    & 47,295     \\
    Converted   & 33,789  & 43.56\%    & 33,789     \\
    Synthetic   & 5,679   & 7.32\%    & 5,679      \\
    \bottomrule
  \end{tabular}
    \caption{Sources of instructions in PLLuM -- Structure of training dataset and total quantity.}
  \label{tab:sft-composition}
\end{table*}

Developing a carefully curated instruction corpus makes it possible to analyse how different instruction types' diversity, quality, and quantity impact the performance of fine-tuned models. In this section, we demonstrate that fine-tuning organically sourced instructions enhances the model’s capabilities, particularly in areas where continued pre-training on textual data and mass distillation from \emph{strong} seemingly multilingual LLMs achieve suboptimal proficiency such as language and culture-specific forms of written communication.

\subsection{Base Model Adaptation and Fine-tuning}\label{subs:lang_adapt}

One of the primary goals of the PLLuM project was to adapt existing base models (see Table \ref{tab:pllum_models}), to better support understanding and generation of native Polish texts. This was partly achieved through (a) continued pre-training of the base models on a corpus of approx. 150 billion tokens, compiled from diverse textual sources, and (b) using a subset of this data to \emph{anneal} the resulting model. At the same time, we observed that certain functional types of texts, which may be particularly important for the intended use of the model, are either underrepresented in the raw pre-training data or, when included, are substandard in terms of style, grammar, and formatting. 

For example, while the pre-training data included Polish e-mails and high-quality style guidelines for e-mail writing, the actual e-mail messages found in the raw corpus data often exhibited non-standard spelling and inconsistent punctuation. To illustrate, normative Polish e-mail style dictates that the second line of a message should begin with a lowercase letter if the first line contains an addressative form followed by a comma, as in:

\begin{quote}
\begin{itshape}
Szanowny Panie,

\hl{c}hciałbym uzyskać informację w sprawie wymiany licznika energii... 
\end{itshape}
\end{quote}

Although this differs from the English convention, where the second line is always capitalized, both capitalization patterns appear in naturally occurring Polish e-mails. Another subtle and frequently ignored prescriptive rule in Polish e-mail writing is the avoidance of a comma between complementary closings and newline signatures, as in:

\begin{quote}
\begin{itshape}
Pozdrawia\hl{m}

Jan Kowalski
\end{itshape}
\end{quote}

Again, this differs from English-language e-mail conventions, where complementary closings are usually separated with a comma from signatures. Although early versions of our SFT models alternated between both conventions when prompted to produce e-mails, we found that a subset of less than 100 high-quality e-mail writing instructions was sufficient to imprint the above-mentioned (and several other) guidelines in the fine-tuned model.
Based on our experience, the need for idiomatic hand-written instructions becomes particularly evident in adapting multi-lingual base models, which were pre-trained mostly on languages other than Polish. We found that special care is required when using fine-tuned LLMs for distilling language-specific instructions. The following is an example of a GPT-4 style Polish email message that illustrates the latter point:

\begin{quote}
	\begin{itshape}
		Szanowny Panie Profesorze,
		\colorbox{yellow}{Mam nadzieję, że ten email zastanie}\\
		\colorbox{yellow}{Pana w dobrym zdrowiu i nastroju.}
		Chciałabym/chciałbym uprzejmie po\-pro\-sić o możliwość umówienia się na \colorbox{yellow}{krótką konsultację} w najbliższą środę o godzinie 11:00. 
		(...)
		Z góry \colorbox{yellow}{dziękuję za poświęcony czas} i roz\-ważenie mojej prośby.
		Z poważaniem\colorbox{yellow}{,}
		Twoje imię i nazwisko
	\end{itshape}
\end{quote}

While the email successfully fulfills the communicative goal of scheduling an appointment (as specified in the prompt), it also contains several instances of negative linguistic transfer from English. Beyond the formatting and punctuation violations mentioned earlier, the opening sentence directly translates a formulaic English email introduction (\textit{I hope this message finds you in good health}), which sounds unidiomatic in Polish. This exemplifies a broader issue in LLM transfer learning: while models are designed to generalize across languages, their ability to transfer knowledge and skills can sometimes manifest as unintended stylistic interference. Transformer-based models tend to transfer stylistic conventions from languages best represented in the pre-training phase, leading to non-idiomatic outputs in the target language. This is similar to the human-like transfer of syntactic and pragmatic constructions from native or otherwise predominant language \cite{selinker1969language}.

\subsection{Alignment}

Model alignment on preference-based datasets, where chosen and rejected response pairs are annotated according to human preferences, aims to teach the model appropriate behaviours, particularly in responding to controversial and potentially harmful prompts.
For the alignment of the PLLuM models, we used a dataset of over 40,000 manually annotated instructions, derived from three distinct annotation methodologies:

\begin{itemize}
\setlength{\itemsep}{0pt}
\setlength{\parskip}{0pt}
 \item Rating-based annotation – each response was assessed according to a predefined metric, with the higher-rated response designated as the preferred (chosen) response.
 \item Ranking-based annotation – four responses were ranked according to response quality.
 \item Dialog-based annotation – annotators engaged in multi-turn interactive conversations with the model, selecting the most appropriate responses.
\end{itemize}

The prompts were primarily created manually and did not overlap directly with PLLuMIC, although they were based on a similar typology, with a strong emphasis on safety-related prompts. Responses were generated by various open models, including the PLLuM ones. In cases where no response met the established criteria, annotators (over 50 different persons in total) provided their own responses.

Our experimental results indicate that, within the scope of this study, the most effective alignment method was the Odds Ratio Preference Optimization (ORPO) algorithm \cite{hong2024orpo}, which integrates alignment with instruction tuning through a specifically designed loss function. While previous research suggests that employing such a loss function obviates the need for SFT, our findings demonstrate that applying ORPO after SFT still yielded superior performance compared to alternative approaches such as KTO \cite{ethayarajh2024}, DPO \cite{rafailov2024}, and PPO \cite{schulman2017}. 

Through alignment training, model safety behaviours improved significantly, enabling our models to proactively address adversarial inputs and provide well-reasoned, defensible explanations, confirmed by red-teaming evaluation results (see \ref{evaluation}). At the same time, we observed the well-documented trade-off between safety and helpfulness—a tendency for models to overly refuse to respond \cite{bai2022}, even in the case of non-adversarial prompts (those that do not contain harmful content or encourage unsafe behaviour). While factuality and linguistic correctness remained mostly consistent with those achieved through SFT, verbosity increased, with models demonstrating a tendency to generate more elaborate responses, even in cases where a more concise answer would have been sufficient. 

At the same time, we also observed negative language transfer at this stage, stemming from the preference for responses generated by models trained predominantly on English-language data. Without additional linguistic adjustments, alignment on the preference-based dataset occasionally resulted in grammatical, lexical, and stylistic inconsistencies reflecting English-language rules, many of which had already been addressed during the SFT phase (e.g., punctuation in emails). This highlights the fact that for both SFT and alignment in low- and mid-resourced languages, manual human annotation, evaluation, and quality assurance are indispensable for maintaining proper language standards.

\subsection{Evaluation}
\label{evaluation}

The linguistic adaptation of the base models listed in Table \ref{tab:pllum_models} was evaluated on the Polish Linguistic and Cultural Competency Benchmark (PLCC)~\cite{dadas2025evaluatingpolishlinguisticcultural}, which consists of 600 questions covering topics such as Polish history, geography, culture, tradition, art, entertainment, grammar, and vocabulary. The answers to the questions are assessed using an IFEval evaluation scheme \cite{zhou_instruction-following_2023}. It is important to note that the PLLuM instruction corpus was developed independently from this benchmark. The base models were fine-tuned for 3 epochs using the AdamW optimizer (weight decay: 0.1) with a learning rate of 1e-5, a cosine scheduler (1\% warmup), and a cumulative batch size of 128 on PLLuM instructions with a maximum sequence length of 16 384 tokens. The loss values were calculated only on the response turn of the instructions. The training was performed in a multi-node configuration\footnote{We used NVIDIA H100 nodes maintained by the Wrocław Centre for
Networking and Supercomputing.} using DeepSpeed ZeRO Stage 3 optimization.

Table \ref{tab:plcc_benchmark} shows the PLCC scores obtained for the different stages of linguistic adaptation. Each group of evaluated models consists of:

\begin{enumerate}
    \item The original reference instruction-following model, e.g. Mistral-Nemo-Instruct-2407,
    \item The reference base model fine-tuned on PLLuMIC, e.g. Mistral-Nemo-2407+PLLuMIC,
    \item A model continually pre-trained on Polish texts and fine-tuned on PLLuMIC, e.g. PLLuM-12B-nc-instruct,
    \item The continually pre-trained model, fine-tuned on PLLuMIC, aligned on PLLuM human preferences, e.g. PLLuM-12B-nc-chat.
\end{enumerate}

Several conclusions emerge from this evaluation. Firstly, the continually pre-trained models consistently outperform their base counterparts across all four architectures. Secondly, fine-tuning on PLLuMIC is only effective for models that have undergone continual pretraining; otherwise, fine-tuning even degrades the model performance. In other words, fine-tuning on Polish instructions requires a model sufficiently primed on Polish data in the pre-training phase. To further examine the relationship between these two training phases, we have conducted additional ablation experiments, described in the Appendix \ref{sec:abl_exp}. Finally, models aligned with human preferences achieve slightly higher benchmark scores than their instruction-fine-tuned predecessors. This might be partly because aligned models usually generate longer responses, which increases their chances of meeting the inclusion criteria of IFEval-style benchmarks. For example, if a model frequently paraphrases or summarizes parts of its responses, it is more likely to include words that match the benchmark criteria, thus improving its overall score.

The general knowledge capabilities (in contrast to the more cultural or linguistic competences) of the models fine-tuned on the PLLuM instruction corpus was also evaluated on the LLMzSzŁ benchmark (see Table~\ref{tab:llmzszl_benchmark}), which is ``a collection of Polish national exams, including both academic and professional tests extracted from the archives of the Polish Central Examination Board'' \cite{jassem2025llmzszl}. Interestingly, the performance of our LLama-PLLuM-70B-chat model, which was fine-tuned on our instructions is only 2.71 points lower than the performance of LLama-3.3-70B-Instruct, which is reported to have been fine-tuned on millions of manually crafted instructions \cite{llama3modelcard}.

\begin{table}

    \centering
    \begin{tabular}{lc}
        \toprule \textbf{Model} &  \textbf{PLCC} $\uparrow$ \\ \midrule
        Mistral-Nemo-Instruct-2407 & 23.00 \\
        Mistral-Nemo-2407+PLLuMIC & 22.33 \\
        PLLuM-12B-nc-instruct & 56.33 \\
        PLLuM-12B-nc-chat & 59.50 \\ \hline
        Mixtral-8x7B-Instruct-v0.1 & 35.33  \\
        Mixtral-8x7B-v0.1+PLLuMIC & 32.17  \\
        PLLuM-8x7B-nc-instruct & 67.17 \\
        PLLuM-8x7B-nc-chat & \textbf{68.17} \\ \hline
        Llama-3.1-8B-Instruct & 22.67 \\
        Llama-3.1-8B+PLLuMIC & 24.67 \\
        Llama-PLLuM-8B-instruct & 58.00  \\
        Llama-PLLuM-8B-chat & 60.67 \\ \hline
        Llama-3.1-70B-Instruct & 47.83 \\ 
        Llama-3.1-70B+PLLuMIC & 38.67 \\ 
        Llama-PLLuM-70B-instruct &  65.17 \\
        Llama-PLLuM-70B-chat &  66.33 \\ \hline
        Qwen-Max & 50.83 \\
        GPT-4 &  59.50 \\
        Grok-2-1212 & 66.00 \\
        DeepSeek-v3 &  69.17 \\
        DeepSeek-R1 & 76.00 \\ 
        O1-2024-12-17 & \textbf{89.17} \\ \bottomrule
    \end{tabular}
    \caption{Linguistic adaptation rate as evaluated on the \href{https://huggingface.co/spaces/sdadas/plcc}{PLCC}: Polish Linguistic and Cultural Competency Benchmark}
    \label{tab:plcc_benchmark}
\end{table}

\begin{table}
    \centering
    \begin{tabular}{lcc}
        \toprule \textbf{Model} &  \textbf{ASR}$\downarrow$ & \textbf{FRR} $\downarrow$ \\ \midrule
        Mistral-Nemo-Instruct-2407 & 21.85 & 0.62 \\
        PLLuM-12B-nc-base & 72.80 & 10.90 \\
        PLLuM-12B-nc-instruct & 77.61 & 0.62 \\
        PLLuM-12B-nc-chat & 1.03 & 3.31 \\ \hline
        Mixtral-8x7B-Instruct-v0.1 & 31.86 & 0.59 \\
        PLLuM-8x7B-nc-base & 74.35 & 6.95 \\
        PLLuM-8x7B-nc-instruct & 70.63 & 0.56 \\
        PLLuM-8x7B-nc-chat & 0.78 & 8.69\\ \hline
        Llama-3.1-8B-Instruct & 19.66 & 0.86 \\
Llama-PLLuM-8B-base & 80.02 & 3.86 \\
        Llama-PLLuM-8B-instruct & 78.60 & 1.2 \\
        Llama-PLLuM-8B-chat & \textbf{0.76} & 5.27 \\ \hline
        Llama-3.1-70B-Instruct & 22.27 & \textbf{0.36} \\
Llama-PLLuM-70B-base & 76.35 & 2.01\\
        Llama-PLLuM-70B-instruct & 70.69 & \textbf{0.36} \\
        Llama-PLLuM-70B-chat & 0.79 & 5.22 
         \\ \bottomrule
    \end{tabular}
      \caption{Red-teaming evaluation results.}
    \label{tab:red-teaming}
\end{table}

Finally, the red-teaming evaluation results of our models are summarized in Table \ref{tab:red-teaming}. The evaluation was conducted on 18,656 harmful prompts for the attack success rate (ASR) metric and 9,724 non-harmful samples for the false-refusal rate (FRR) metric \cite{krasnodebska2025rainbow}. Both datasets cover 14 hazard categories defined by the Llama-Guard taxonomy \cite{inan2023}. Additionally, they were generated using 10 different attack styles inspired by the "Rainbow Teaming framework" \cite{samvelyan2024}. For the ASR, the Llama-Guard model was utilized to assess the percentage of unsafe responses, whereas for the FRR, we prompted one of our trained models to obtain the proportion of refusals to benign queries. In general, the PLLuM models fine-tuned on instructions are characterized by a relatively higher ASR and a lower FRR than their derivatives aligned on human preferences.

\begin{table}
    \centering
    \begin{tabular}{lc}
        \toprule \textbf{Model} &  \textbf{LLMzSzŁ} $\uparrow$ \\ \midrule
        Llama-PLLuM-8B-chat & 47.68 \\
        PLLuM-12B-nc-chat & 53.40 \\
        PLLuM-8x7B-nc-chat & 60.52 \\
        Llama-PLLuM-70B-chat & \textbf{64.42} \\ \hline 
        Meta-Llama-3.1-8B-Instruct & 47.41 \\
        Mixtral-8x7B-Instruct-v0.1 & 49.46 \\
        Bielik-11B-v2.1-Instruct & 57.52 \\
        Llama-3.3-70B-Instruct & \textbf{67.13} \\ \bottomrule
    \end{tabular}
        \caption{Academic performance as evaluated on \href{https://huggingface.co/spaces/amu-cai/LLMZSZL_Leaderboard}{LLMzSzŁ}: a comprehensive LLM benchmark for Polish}
    \label{tab:llmzszl_benchmark}
\end{table}

\section{PLLuMIC Public Sample}

Apart from evaluating the impact of the different phases of model training on its linguistic adaptation, we release a representative subset of the organic PLLuM instruction corpus. Overall the first release of the dataset contains a total of 1278 human-authored instructions, spanning across 12 types, 126 subtypes and 34 topics. Substantial effort has been made to ensure a wide diversity and high quality, both reflected in each data sample. Appendix \ref{sec:appendix_organic_typology} details the subset's typology.

\section{Conclusions}

We believe that our description of the PLLuM Instruction Corpus along with its public subset can be used to design and complement manual and automated annotation work in other LLM projects. Thanks to iterative instruction corpus development and continual evaluation we established that effective fine-tuning on language-specific instructions requires models to first undergo sufficient continual pre-training on the target language. We also identified several cases of negative linguistic transfer, where conventions from dominant languages (particularly English) can interfere with idiomatic text generation in the target language. Such interference may occur both in the process during fine-tuning and alignment on synthetic instructions. This highlights the need for high-quality, language-specific organic instructions in linguistically adapted LLMs.

\section{Availability}
\label{sec:availability}

The manually annotated sample of PLLuMIC can be accessed at \url{https://huggingface.co/datasets/pelcra/PLLuMIC}. We are planning to release its synthetic extension (PLLuMIC-syn-ext) separately.


\section{Acknowledgments}
\label{sec:acknowledgments}
The work reported in this paper was funded by several grants: 

\begin{itemize}
\item The continued pre-training of the 8B, 12B, and 70B models reported in Table \ref{tab:plcc_benchmark}, most of their fine-tuning and alignment were performed on the WCSS HPC infrastructure as part of an earmarked grant (1/WI/DBiI/2023) from the Polish Ministry of Digital Affairs. 
\item The continued pre-training of variants of the 8x7B and 12B models reported above were performed on the ACC Cyfronet AGH infrastructure under a grant no. PLG/2024/017788. 
\item The first edition of the PLLuMIC subset released with this paper was developed after the completion of the PLLuM project and supported by the grant CLARIN-BIZ-bis (FENG.02.04-IP.04-0004/24).
\end{itemize}

\bibliography{bib/custom}

\appendix




\section{Availability of Instruction Datasets}
\label{sec:appendix_transparency}

\subsection{Selected LLMs and their source datasets}

\begin{table*}[ht]
  \centering
  \caption{Availability of instruction datasets for selected open LLMs. We characterize LLMs especially based on the availability of information concerning the annotation process and synthetic data generation (SDG). Ideally, we would expect the final instruction mix used in SFT to be fully documented (e.g. exact proportions of each instruction type).}
  \begin{tabular}{p{0.3\linewidth}  p{0.3\linewidth}  p{0.3\linewidth}}
    \hline
    \textbf{Model} & \textbf{Instructions}  & \textbf{Documentation} \\
    \hline
    \verb|LLaMA 3.1|  \cite{llama31models}    & Unavailable  & SDG \&  annotation process explained     \\
    \verb|Mixtral|    \cite{jiang_mixtral_2024}      & Unavailable      & \cite{jiang2023mistral7b} points to \say{instruction datasets publicly available on the Hugging Face repository}      \\
    \verb|Falcon3|  \cite{Falcon3family}       & Unavailable      & None     \\
    \verb|Gemma| \cite{team_gemma_2024-1}  & Unavailable      & High-level desc. of SDG process     \\
    \verb|Nemotron| \cite{nvidia_nemotron-4_2024} & As input for SDG      & SDG process explained     \\
    \verb|Snowflake-arctic| \cite{snowflake_artic_blog}  & Unavailable      & None     \\
    \verb|Phi-3.5/4| \cite{abdin_phi-4_2024} & Unavailable      & High-level desc. of SDG process     \\
    \verb|Dolly| \cite{DatabricksBlog2023DollyV2}  & Available      & Annotation process explained     \\
    \verb|OpenChat| \cite{openchatpaper}   & Unavailable     & Low     \\
    \verb|Qwen 2.5| \cite{qwen2025qwen25technicalreport}   & Unavailable      & Low     \\
    \verb|DeepSeek|  \cite{deepseek-ai_deepseek-v3_2024}  & Unavailable      & SDG process explained     \\
    \verb|OLMO| \cite{olmopaper}  & Available    & Inventory of component datasets    \\
  \bottomrule
  \end{tabular}
  \label{tab:open-weight-models}
  \end{table*}

Table \ref{tab:open-weight-models} summarizes the availability status of instruction datasets for a number of open-weight models. More specifically, Llama 3.1 \cite{llama31models} offers a general description of the data preparation process, including the sampling of synthetic and human-annotated instructions, but does not share the data itself.\\
OpenChat utilizes the acclaimed ShareGPT\footnote{No longer available online.} dataset of prompts and responses generated by OpenAIs GPT-3.5 and GPT-4. The accompanying paper \cite{openchatpaper} offers a condensed analysis of the data distribution and quality.\\
Qwen 2.5 \cite{qwen2025qwen25technicalreport} uses various datasets synthesized according to their guidelines during the post pre-training phase \cite{quan2024languagemodelsselflengthengenerate,dong2024selfplayexecutionfeedbackimproving}, either sampling existing datasets (e.g. \cite{chai2024mcevalmassivelymultilingualcode}) or using web-scrapped data as input.\\Qwen2 \cite{yang2024qwen2technicalreport} gives out more details concerning the human-annotation process in the data generation process. However, no ready-to-use data was published alongside these models.\\
OLMO's \cite{olmopaper} fine-tuning is based on the Tulu2 dataset \cite{ivison2023camelschangingclimateenhancing}, recently developed into Tulu3 \cite{lambert2024tulu3}. Tulu2 consists of publicly available datasets such as FLAN \cite{flan2022}, No Robots \cite{no_robots} and WildChat \cite{zhao2024wildchat1mchatgptinteraction}.\\
The authors of Mixtral 8x7B \cite{jiang_mixtral_2024} offer no description of the employed instruction data, while the publication of Mistral 7B \cite{jiang2023mistral7b} points to loosely defined ``instruction datasets
publicly available on the Hugging Face repository. No analysis can be found in either of the papers.\\
Falcon \cite{Falcon3family, almazrouei_falcon_2023} is predominantly focused on pre-training but its various \textit{-instruct} versions utilize mainly Baize \cite{xu2023baizeopensourcechatmodel} dataset, sourcing also from other GPT4-based online data repositories, such as GPT4All \cite{gpt4all} or GPTeacher.\\
Gemma \cite{team_gemma_2024-1} uses an undisclosed teacher-model to generate answers for synthetic and human-made prompts, joining it with a \say{mixture of internal and external public data} but offers no insight as for the composition of these datasets.\\
Microsoft's Phi model \cite{abdin_phi-4_2024} uses undisclosed publicly available datasets to generate synthetic responses for supervised fine-tuning (SFT) and gives no detailed description of their contents.\\
In contrast, Dolly \cite{DatabricksBlog2023DollyV2} use their open-source dolly-bricks-15k dataset comprising 15k human-generated prompt-response pairs inspired by InstructGPT \cite{ouyang2022traininglanguagemodelsfollow}, accompanied by extensive documentation, including annotation guidelines.\\
For their V3 model \cite{deepseek-ai_deepseek-v3_2024}, Deepseek uses other iterations of models such as V2.5 \cite {deepseekv2} or R1 as ``expert models'' to generate instruction data from scratch.\\
Nvidia's Nemotron \cite{nvidia_nemotron-4_2024} relies on their own Helpsteer2 dataset \cite{wang2024helpsteer2opensourcedatasettraining} and Mixtral-8x7b's abilities to generate synthetic instruction data. The data generation process is documented, but only the seed dataset is available. 
\clearpage
\section{Stand-alone Instruction Datasets}

\subsection{Transparency and Representativeness}

Several instruction datasets available as open stand-alone collections have also been used in more experimental LLM research projects. One of the early large-scale resources of instructions is the original FLAN \cite{flan2022} dataset, followed by the FLAN Collection \cite{longpre2023flan} dataset published by Google Research.\\
The OpenOrca\footnote{\mbox{See\,also\,its\,filtered\,version\,--\,SlimOrca\,\cite{SlimOrca}.}} \cite{OpenOrca} dataset complemented FLAN with explanation traces and step-by-step thought processes from GPT-3.5 \cite{openai2023gpt35} and GPT-4 \cite{peng2023instructiontuninggpt4}.\\
Other frequently utilized datasets include  Databrick's Dolly \cite{DatabricksBlog2023DollyV2} 15k and  LIMA \cite{zhou_lima_2023} 1k dataset both consisting of curated and hand-written examples, LMSYS-Chat-1M \cite{zheng2023lmsyschat1m} with 1 million human-LLM conversations, WebInstruct \cite{yue2024mammoth2} with 10 million instruction pairs harvested and refined from the web, UltraChat \cite{ding2023enhancingchatlanguagemodels} comprising 1.5 million multi-turn dialogues generated by ChatGPT \cite{openai2022chatgpt} from C4 data, SelfInstruct \cite{selfinstruct} containing 52k synthetic instructions bootstrapped from 175 hand-written examples by GPT-3 \cite{ouyang2022traininglanguagemodelsfollow}. Similarly, the Stanford Alpaca dataset \cite{alpaca} contains 52k instructions created with OpenAI's text-davinci-003 model \cite{openai2023gpt35} and it was subsequently reconstructed with GPT-4 and extended to 110k items \cite{instructionwild}.The HelpSteer datasets \cite{wang2023helpsteer, wang2024helpsteer2opensourcedatasettraining} contain prompts sourced from ShareGPT dataset and answers generated by Nemotron and Mixtral-8x7B, later augmented by human annotators. This is further summarized in Table \ref{tab:open-sft-datasets}

\begin{table*}[ht]
  \centering
  \caption{Availability of stand-alone instruction datasets.}
  \begin{tabular}{p{0.3\linewidth}  p{0.6\linewidth}}
    \hline
    \textbf{Dataset} & \textbf{Contents}  \\
    \hline
    \verb|FLAN Collection|  \cite{longpre2023flan}     & Collection of Google datasets, e.g. original FLAN \cite{flan2022}, P3/T0 \cite{sanh2022multitaskpromptedtrainingenables},  Natural Instructions \cite{wang2022flan2022component}  \\
    \verb|OpenOrca| \cite{OpenOrca}      & Augmentation of FLAN Collection datasets with GPT-3.5 \cite{openai2023gpt35} and GPT-4 \cite{peng2023instructiontuninggpt4} completions    \\
    \verb|Dolly|  \cite{DatabricksBlog2023DollyV2}  & Hand-written instructions prepared according to InstructGPT \cite{ouyang2022traininglanguagemodelsfollow} guidelines     \\
    \verb|LIMA|  \cite{zhou_lima_2023}  & Hand-written instructions curated from online Q\&A forums     \\
    \verb|LMSYS-Chat-1M| \cite{zheng2023lmsyschat1m}       & Human-AI conversations with 25 different LLMs  \\
    \verb|UltraChat|   \cite{ding2023enhancingchatlanguagemodels}       & Synthethic multi-turn dialogues generated with ChatGPT \cite{openai2022chatgpt}   \\
    \verb|Selfinstruct|  \cite{selfinstruct}                 &  Synthetic instructions bootstrapped from 175 hand-written examples by GPT3 \cite{ouyang2022traininglanguagemodelsfollow}.      \\
    \verb|HelpSteer|  \cite{wang2023helpsteer, wang2024helpsteer2opensourcedatasettraining}     & Prompts sourced from ShareGPT and answers generated by Nemotron and Mixtral-8x7B, later augmented by human annotators.   \\
    \verb|Open-Playtyps| \cite{lee2024platypus}      &  Subsets of 11 specific-domain datasets such as MATH \cite{hendrycksmath2021}, PRM800K \cite{lightman2023lets}, ScienceQA \cite{lu2022learn} or TheoremQA \cite{chen2023theoremqa}.     \\
    \verb|Tulu| \cite{tulu3dataset}   &   Composition of domain-specific data such as Tabe-GPT \cite{li2023tablegpttabletunedgptdiverse}, SCRiFF \cite{wadden2024sciriffresourceenhancelanguage} or coconot \cite{brahman2024artsayingnocontextual}   \\
    \verb|Open Hermes| \cite{OpenHermes2.5}     & Subsets from Open-Playtypus and SlimOrca, but also from ShareGPT and MetaMathQA \cite{yu2024metamathbootstrapmathematicalquestions}    \\
    \verb|Infinity Instruct|  \cite{zhang2024inifinitymath, zhao2024iidoptimizinginstructionlearning} & Samples from OpenHermes, FLAN, UltraChat, Dolly Dataset complemented by DEITA \cite{liu2024what} and CodeFeedback \cite{zheng2025opencodeinterpreterintegratingcodegeneration}    \\
  \bottomrule
  \end{tabular}
  \label{tab:open-sft-datasets}
  \end{table*}


More focused, special-domain collections have also been released. For example, Open-Playtypus \cite{lee2024platypus} comprises subsets of 11 specific-domain datasets such as MATH \cite{hendrycksmath2021}, PRM800K \cite{lightman2023lets}, ScienceQA \cite{lu2022learn} or TheoremQA \cite{chen2023theoremqa} curated into a sample of 25k thematically versatile question-answer pairs. Similarly, AllenAI's Tulu \cite{tulu3dataset}  contains domain-specific data such as Tabe-GPT \cite{li2023tablegpttabletunedgptdiverse}, SCRiFF \cite{wadden2024sciriffresourceenhancelanguage} (54 scientific literature understanding tasks) or coconot \cite{brahman2024artsayingnocontextual} with 13k non-compliance examples. Multiple collection datasets overlap each other: OpenHermes \cite{OpenHermes2.5} contains subsets from Open-Playtypus and SlimOrca, but also from ShareGPT and MetaMathQA \cite{yu2024metamathbootstrapmathematicalquestions}. InfinityInstruct \cite{zhang2024inifinitymath, zhao2024iidoptimizinginstructionlearning} contains samples from OpenHermes, FLAN, UltraChat, Dolly Dataset complemented by DEITA \cite{liu2024what} and CodeFeedback \cite{zheng2025opencodeinterpreterintegratingcodegeneration}. Recently, with the advent of reasoning abilities in LLMs, datasets such as Bespoke-Stratos \cite{bespoke_stratos}, OpenThoughts \cite{openthoughts} or Magpie-Align \cite{xu2024magpie} - covering chain-of-thought traces for math, science, and puzzle-solving.

Various other datasets, often lacking publication or licensing information, can be found in large aggregated instruction corpora. Examples of such meta-sets include DialogStudio \cite{zhang2023dialogstudio}, LlamaFactory \cite{zheng2024llamafactory}, and AlpacaCoT \cite{alpaca-cot}, which serve as comprehensive frameworks for both dataset curation and LLM training.
\section{Summary of Annotation Guidelines}
\label{sec:appendix_ann_guide}
\textbf{Context} 
Fine-tuning a large language model requires a comprehensive and diverse dataset of instructions. The annotation task involves the manual creation of two-element instructions, consisting of a prompt and a correct response. In the case of multi-turn instructions, each turn is represented by a single prompt-response pair. Additional elements, such as argumentation, context, or keywords, are included only for specific subtasks.
The annotation process is carried out using dedicated annotation sheets, with each annotator assigned sheets tailored to different instruction types. The typology of instructions is aligned with the PLLuMIC framework.

\subsection{Quality Control Measures}
\label{subsec:quality_control}
To guarantee the highest possible quality of the annotated samples, we have introduced multiple quality assurance steps and provide comprehensive details on annotator qualifications and quality metrics.

All annotators (over 50 in total) were hired on an employment contract. They were all university graduates, with at least a bachelor's or master's degree in linguistics or other humanities with the exception of technical instructions annotators who had a university degree in computer science. All of the super-annotators had a PhD degree.

We did not use inter-annotator agreement scores as we feel that they are not directly suitable for most of the generative and extractive tasks covered in the LLM instruction dataset (e.g. email writing, multiturn dialogs etc.). Agreement scores calculations are typically used in labeling or rating dataset development scenarios with deterministic outcomes/ answers. Instead, we have implemented a number of other measures to maximize consistency and high quality of the instruction dataset:
\begin{itemize}
\item Detailed annotation guidelines were developed and adjusted throughout the project (see the remaining sections of this Appendix \ref{sec:appendix_ann_guide}).
\item A four week training period for new annotators to master annotation guidelines and standards.
\item Weekly team meetings provided ongoing coordination, allowing annotators to discuss current and new tasks and maintain consistency.
\item A quality assurance process where an experienced super-annotator reviewed all instructions and provided targeted feedback to address any problematic elements.
\end{itemize}

\subsection{Single Turn Instructions}
\textbf{General guidelines}
\begin{itemize}
\item Linguistic accuracy is crucial. Responses to prompts must be written in correct, high-register Polish free of typos, punctuation errors and grammatical mistakes, with generally high stylistic quality. 
\item In prompts, grammatical gender should be varied when necessary — most prompts are written using impersonal, gender-neutral forms, but masculine and feminine pronouns and inflections should be used interchangeably when required. Model responses and argumentation should preferably be structured in a way that does not reveal gender, but if it cannot be avoided (e.g. in role-playing tasks or identity questions), the model by default uses the masculine gender because the Polish word 'model' is a masculine noun taking masculine inflectional endings. Nevertheless, the model switches to feminine forms when asked to change the forms or when a given role requires it.
\item Questions may be informal, but model responses should always be formal unless the generative task requires an informal style (e.g., in an email to a close friend or a social media post). 
\item Responses must be carefully formatted according to separate detailed formatting guidelines, which include punctuation rules for bullet points and labeled lists, text structure, spacing, bold and italic text, headings, indentation, emoticons, citations, code blocks, mathematical formulas, and tables. Markdown formatting is used by default, while LaTeX is combined with Markdown for mathematical expressions.
\item Categorical statements should be avoided unless the model presents factual knowledge that cannot be disputed. When discussing rules and ethical dilemmas, instead of absolute claims, the model should lean towards more hedged phrases such as \textit{in most societies it is not accepted}, \textit{one should not}, \textit{it is better not to}. In contrast, unqualified uses of \textit{you cannot} or \textit{it is not allowed to} are avoided. Instead, such statements should be supported by a knowledge source, e.g., \textit{According to the regulations from [date], it is not allowed to...}
\item For questions requiring subjective opinions or value judgments, the model's responses should remain neutral. For instance, when asked \textit{Is coffee better than tea?}, the expected response might be: \textit{It depends on individual preferences. Some people cannot imagine life without coffee, while others simply dislike it. Similarly, tea is also widely enjoyed, and both beverages are popular in Poland.}
\item Single-turn instructions should be understandable without additional context. For example, avoid questions like \textit{Does the same price list apply when issuing a second permit as for the first one?} — linked instruction series are developed as a separate task (see \textit{Multiple turn instructions}).
\end{itemize}
\textbf{Localization of English-Language Instructions}
\begin{itemize}
\item When classifying an instruction as an adaptation (significantly altered version) or a translation (a fairly close rendering), the key criterion is whether the prompt has been modified. Simply improving or expanding the response does not qualify as an adaptation.
\item Whenever an example contains minimal argumentation, we should expand on it to help guide the model in associating knowledge with relevant topics.
\item We freely substitute locations and people, modify contexts, and create instructions embedded in Polish culture, history, and everyday life.
\item Literal translations or translation loans from English must be avoided. Instead, we should look for natural Polish equivalents.
\item For yes/no questions, we generally operate with two types of statements: factual claims (e.g., \textit{Dogs are mammals}) and hypothetical scenarios (e.g., \textit{A dog came to a shop to buy some carrots}). In the former case, we ask whether a given statement is true or factually accurate. In the latter case, we ask whether the statement makes sense or describes a likely situation.
\end{itemize}
\textbf{Knowledge-driven (QA)}
\begin{itemize}
\item For domain-specific instructions, we ensure a variety of questions, and, when addressing the same topic, we try to rephrase subsequent questions to avoid repeating the same pattern.
\item In open-ended questions, the argumentation often mirrors the response. In such cases, argumentation may be omitted.
\end{itemize}
\textbf{Extraction}
\begin{itemize}
\item In instruction representing this type, the prompt must consist of a text excerpt followed by a question related to the text. At the same time, the response should be very specific and concise, followed by a short fragment of the text containing the answer. The fragment must be introduced by a statement explaining that the answer to the question may be found in this particular part of the text. Text excerpts for these instructions are sourced independently from Wikipedia.
\item The response may involve inference (the answer does not have to be explicitly stated in the text).
\item If the answer is spread across two or more separate fragments within the text, they can be combined in the response.
\end{itemize}
\textbf{Generation}
\begin{itemize}
\item The response must not be directly copied from any source (it can be inspired by various sources, but these must be thoroughly paraphrased).
\item If the prompt does not explicitly suggest it, we ensure that the response does not introduce new facts that were not included in the prompt.
\item In prompts, we can provide fictional personal data (which should be fairly ordinary). If the prompt lacks necessary details that should be included in the response, we use placeholders, e.g., [phone number], [email address].
\item The texts used for processing (e.g., paraphrasing or style modification) should be sourced from the public domain (e.g., Wikinews). The prompt should contain the text or its fragment (for paraphrasing and style changes, it should be at least five sentences; for simplifications, 1–2 paragraphs; for summaries, 200–300 words).
\item Responses to requests for formal text generation should be neatly formatted, including all necessary formalities (date, location, sender's address, etc.).
\item We avoid socially sensitive topics (crime, alcohol, drugs, violence), erotic content, and themes that could be offensive to any minority group.
\item For prompts requesting creation of a test or quiz, the response should include at least five test questions along with an answer key.
\item For prompts requesting lists of ideas or recommendations, the response should contain a short introduction followed by a list (preferably with each item accompanied by a brief explanation or justification).
\item For prompts requesting a review, the response should be a collection of facts (e.g., a summary of the plot, description of the object, its popularity backed by awards and sales figures) rather than a categorical evaluation.
\item For prompts requesting a comparison of two objects, products, countries, people, animals, etc., the response should be an objective comparison based on factual differences. Evaluative statements should be avoided. The response should begin with a brief introduction and end with a concluding summary of the comparison.
\item Prompts asking the model to generate a short conversation should include additional details such as the topic, conversation style, etc. The response should be a short dialogue between X and Y, with each line starting with the character’s name followed by a colon. Conversations should be created in diverse styles.
\end{itemize}
\textbf{Formatting \& visualization}
\begin{itemize}
\item Transformations may involve retrieving responses from the model's knowledge base or context. Previously developed instructions of other types can be used as the basis. If external sources are used, they must be open, such as Wikipedia.
\item Transformations include modifying the paragraph structure, adding headers, creating and formatting lists, inserting content at specific locations, adding introductions or summaries, formatting individual words, changing capitalization, modifying punctuation, introducing bolding and italics, and creating or modifying tables.
\item If the prompt does not specify the number of list elements, the response should clarify this: the model should start by explaining how many items it will include in the list or use a phrase like “a few.”
\item The response should be appropriately formatted according to Markdown guidelines, depending on the content.
\item Diagrams, charts, graphs, and other visualizations should be created using Mermaid, a tool that renders Markdown-inspired text definitions to generate and modify diagrams dynamically. Prompts may include syntactic tree diagrams, time series, genealogical charts, database schemas, or class diagrams.
\end{itemize}
\textbf{Data manipulation}
\begin{itemize}
\item Transformations may involve providing responses in a specified format or processing statistical data, including demographic, economic, administrative, geographic, and textual data. Data for processing should be high-quality and sourced from open repositories. Previously developed instructions of other types may also be used as the basis.
\item Transformations involve returning responses in XML or JSON format. Suggested transformations include converting natural language data and lists into JSON/XML, standardizing inconsistent tabular data into JSON/XML, converting JSON to XML and vice versa, filtering, modifying, adding, and deleting keys, renaming keys, and altering nesting structures.
\item The XML or JSON provided in an exemplary response can be generated automatically but must be validated.
\end{itemize}
\textbf{Programming}
\begin{itemize}
\item Instructions can be created for various programming languages.
\item Prompts may include requests for code to solve a given problem or task, code review, debugging, or generating correct code. Additionally, we can ask about specific functionalities or knowledge related to a programming language.
\item We can use responses from the Mixtral-8-22B-Instruct-v0.1 model as a reference, but they must be thoroughly verified for technical accuracy and linguistic correctness.
\item Before inserting code into the annotation sheet, it should be formatted in an appropriate editor. Code blocks should be marked using Markdown syntax, specifying the programming language (e.g., python, c++).
\item The model’s response may, but does not have to, end with a concluding sentence. Each time, we should assess whether it is necessary.
\end{itemize}
\textbf{Translation}
\begin{itemize}
\item Prompts may involve various tasks including: translation of a given text, identifying translation errors, pairing corresponding sentences (translating into another language while adapting to a given context), detecting incorrect translations (with indications of where the translation deviates from the original), completing a task in language A while providing input in language B, generating questions in language A for a text in language B, generating parallel texts in two languages, and extracting named entities (NER) for comparison.
\end{itemize}

\subsection{Multi-turn instructions}
\textbf{General Guidelines}
\begin{itemize}
\item Dialogues can vary in length (from two question-answer pairs to longer conversations). It is best to diversify them by creating short and relatively long dialogues.
\item There are no content restrictions as long as the dialogues do not involve controversial or offensive topics. Writing about subjects you are knowledgeable about and that do not require extensive research is encouraged.
\item As a user, ask follow-up questions about previous responses. It is beneficial to ask the model to elaborate, clarify, correct, or modify its prior response.
\item Context shifts within the same dialogue are allowed; you can request multiple unrelated things. Moreover, returning to an earlier topic is welcome (e.g., discussing topic A, switching to topic B, and returning to topic A).
\item Prompts should have varied styles. Correct grammar, neutrality, and politeness are required only in the model’s responses, while user prompts can have different tones and styles.
\item When responding as a language model, keep answers concise and precise, while ensuring they fully address the prompt without unnecessary details.
\item For factual responses, use publicly available sources but do not cite them in the response.
\item Avoid direct translations of English discourse markers; use natural expressions in the target language.
\item System prompts can define the model’s response style for the entire conversation.
\item Each prompt-response pair in the dialogue should be categorized into one of the following interaction types:
\begin{itemize}
\item role-play – The user asks the model to take on a specific role or character.
\item generative – The user requests text generation.
\item extractive – The user provides a text fragment and asks the model to process or modify it.
\item question-answer – Standard question-and-answer exchanges that do not fit the above categories.
\end{itemize}
\item If a turn does not fit any of the above-mentioned categories, do not label it.
\end{itemize}
\textbf{Adapting English-Language Instructions}
\begin{itemize}
\item Treat the original dialogue as an inspiration rather than a strict template. Focus more on conversation structure and user prompt structure than the exact content.
\item Feel free to add original prompts to enrich the dialogue.
\item Shorten original dialogues where possible, however if the original has only 2-3 turns, keep it unchanged.
\item Regardless of length, preserve its original structure as much as possible.
\item If the dialogue covers a general topic, stay closer to the original content.
\item If the dialogue is highly specific (e.g., deeply rooted in the Anglo-Saxon culture), apply localization in addition to adaptation.
\begin{itemize}
\item Formal localization includes adjusting dates, addresses, and abbreviations to Polish conventions.
\item Cultural localization involves modifying references, scenarios, and social elements to be more relevant to Polish-speaking users.
\end{itemize}
\item If a user prompt includes pasted text for processing, use open-license sources if you cannot create original content.
\end{itemize}
\textbf{Creating dialogues from scratch}
\begin{itemize}
\item If struggling with inspiration, refer to:
\begin{itemize}
\item Pre-made datasets of random question-answer pairs (English).
\item Random conversations (Polish).
\item Example categorized dialogues (various types).
\item Your past instructions (original or adapted).
\end{itemize}
\item Similarly to single-turn instructions, dialogues fall into the following categories:
\begin{itemize}
\item Generative dialogues
\item Extractive dialogues
\item Role-play dialogues
\item QA dialogues
\item Mixed dialogues (containing multiple prompt types).
\end{itemize}
\item Mixed dialogues are common and combine different prompt types (see the reference sheet).
\item Another frequent pattern involves chain-of-thought dialogues, which explore a single main idea in various ways. For examples of all dialogue types, refer to the separate reference sheet.
\end{itemize}

\section{PLLuMIC Typology}
\label{sec:appendix_typology}


\subsection{Manual Instructions}
\label{sec:appendix_organic_typology}

\begin{table}
  \centering
  \begin{tabular}{lc}
    \toprule
    \textbf{Instruction Type} & \textbf{Quantity} \\
    \midrule
    \verb|Adversarial| & 125 \\
    \verb|CoT| & 50 \\
    \verb|Data manipulation| & 88 \\
    \verb|Dialog| & 124 \\
    \verb|Extraction| & 71 \\
    \verb|Formatting| & 87 \\
    \verb|Generation| & 392 \\
    \verb|Identity| & 68 \\
    \verb|Knowledge (QA)| & 80 \\
    \verb|NLP| & 102 \\
    \verb|Programming| & 30 \\
    \verb|Translation| & 61 \\
    \bottomrule
  \end{tabular}
  \caption{Type distribution of organic PLLuMIC}
  \label{tab:pllumic_type_division}
\end{table}

\begin{figure*}[t]
    \centering
    \includegraphics[width=\textwidth, page=1, trim=200 200 200 200, clip]{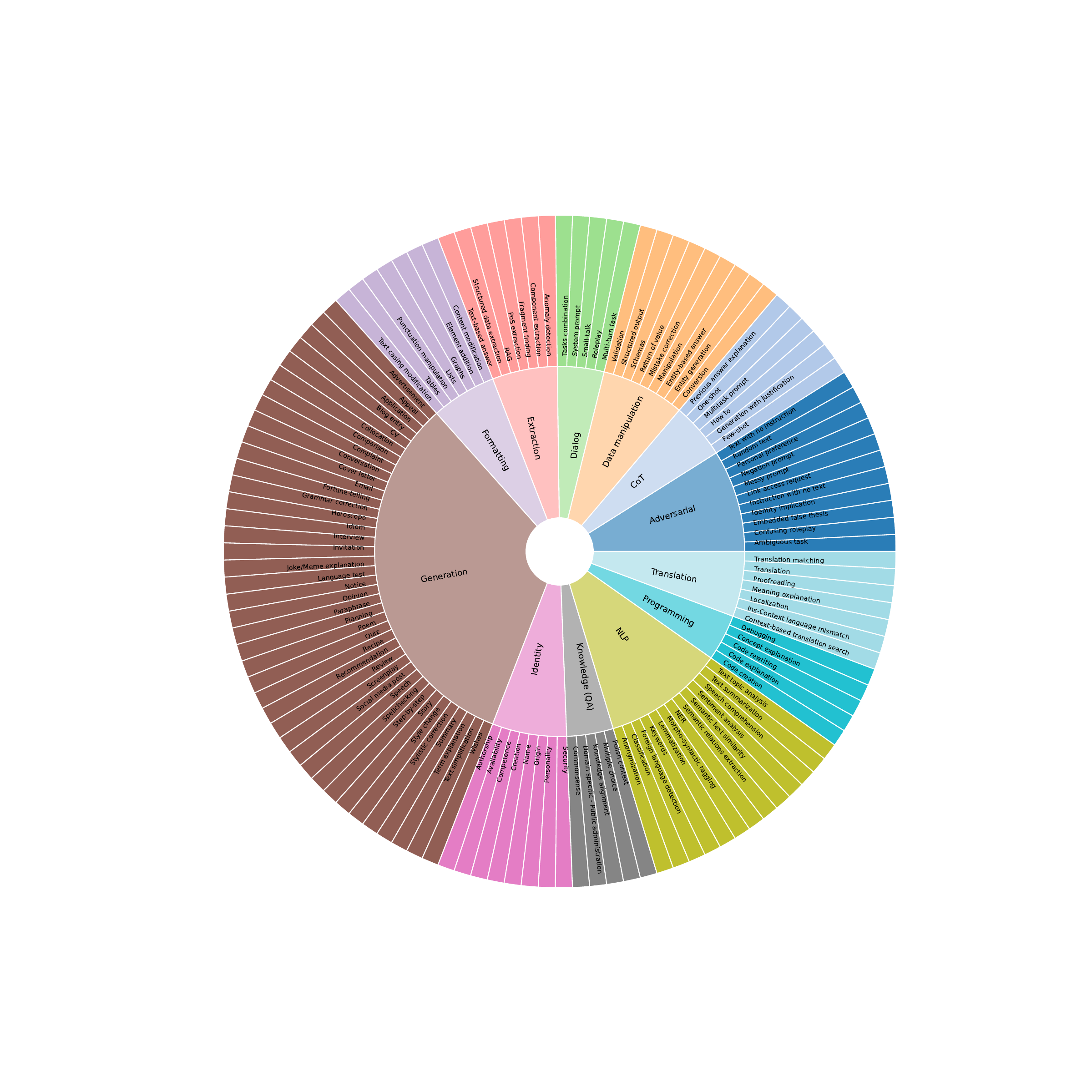}
    \caption{The typology of manual instructions.}
    \label{sec:piechart_typology_vis}
\end{figure*}

The following subsections provide a detailed description of the main functional categories included in the released dataset. The last subsection (\ref{sec:appendix_thematic_categorization}) provides an additional thematic division of the samples. For each individual subtype and topic, the corresponding number of instructions that include it is provided.

The main type distribution is presented in Table \ref{tab:pllumic_type_division}.

\subsubsection{Knowledge-driven (QA)}
\emph{Knowledge-driven} instructions are generally designed to reinforce the factual knowledge representation of the instruction-following model, aligning it with information acquired during the pre-training phase. Since some of them incorporate authentic text samples, they may also strengthen the command of different languages, styles, and registers. 

The QA subset of PLLuMIC comprises the following subtypes:

\begin{itemize}
\setlength{\itemsep}{0pt}
\setlength{\parskip}{0pt}
  \item Common sense (12)
  \item Domain specific - Public administration (12)
  \item Knowledge alignment (29)
  \item Multiple choice (12)
  \item Polish context (15)
\end{itemize}

\subsubsection{Generation}
Instructions classified as \emph{Generation} expose the model to various generative capabilities and formulaic patterns, enabling it to accurately interpret user queries and apply adequate scenarios. The subtypes are very diverse and include a wide range of possible applications, from very formal (e.g. Application, Notice, Complaint) to very informal (e.g. Blog entry, Horoscope, Social media post) and comprising both creation on the basis of a few keywords or just a topic indicated by the user (e.g. Poem, Recipe, Story, Screenplay) and text-based operations (e.g. Spellchecking, Paraphrase, Style change, Text simplification):

\begin{itemize}
\setlength{\itemsep}{0pt}
\setlength{\parskip}{0pt}
  \item Advertisement (13)
  \item Appeal (9)
  \item Application (9)
  \item Blog entry (7)
  \item Collocations (15)
  \item Complaint (8)
  \item Comparison (8)
  \item Conversation (9)
  \item Cover letter (9)
  \item CV (9)
  \item Email (19)
  \item Fortune-telling (9)
  \item Grammar correction (9)
  \item Horoscope (9)
  \item Idiom (14)
  \item Interview (8)
  \item Invitation (9)
  \item Joke/Meme explanation (8)
  \item Language test (11)
  \item Notice (7)
  \item Opinion (10)
  \item Paraphrase (10)
  \item Planning (9)
  \item Poem (10)
  \item Quiz (14)
  \item Recipe (12)
  \item Recommendation (11)
  \item Review (7)
  \item Screenplay (8)
  \item Social media post (12)
  \item Speech (6)
  \item Spellchecking (8)
  \item Step-by-step (11)
  \item Story (11)
  \item Style change (17)
  \item Stylistic correction (7)
  \item Summary (11)
  \item Term explanation (21)
  \item Text simplification (11)
  \item Wishes (10)
\end{itemize}

\subsubsection{Extraction}
Instructions belonging to the \emph{Extraction} type target context-based operations, including text analysis, context-sensitive answer formulation, fragment extraction, and retrieval-augmented generation (RAG) process components. The subtypes include:

\begin{itemize}
\setlength{\itemsep}{0pt}
\setlength{\parskip}{0pt}
  \item Anomaly detection (7)
  \item Component extraction (11)
  \item Fragment finding (15)
  \item PoS extraction (13)
  \item RAG (8)
  \item Structured data extraction (10)
  \item Text-based answer (13)
\end{itemize}

\subsubsection{NLP}
\emph{NLP} instructions enable the model to effectively perform various natural language processing tasks, including classification, named entity recognition, or keyword tagging, with the following subtypes:

\begin{itemize}
\setlength{\itemsep}{0pt}
\setlength{\parskip}{0pt}
  \item Anonymization (8)
  \item Classification (9)
  \item Foreign language detection (7)
  \item Keywords (9)
  \item Lemmatization (7)
  \item Morpho-syntactic tagging (8)
  \item NER (11)
  \item Semantic relations extraction (9)
  \item Semantic text similarity (6)
  \item Sentiment analysis (10)
  \item Text summarization (9)
  \item Text topic analysis (10)
\end{itemize}

Additionally, we include \emph{Speech comprehension} instructions that are intended to enhance the ability of LLMs to process real-world speech scenarios. Answering questions about noisy utterances requires common-sense reasoning and selective processing, particularly the competence to comprehend semantically relevant content while disregarding speech-specific elements, such as substitutions, reformulations, and false starts. The contextual utterances are intentionally selected from DiaBiz \cite{pezik-etal-2022-diabiz} to include \textsl{reparandum} or \textsl{restart} phenomena. The open-ended questions and their answers focus on the reformulated or restarted segments of these utterances:

\begin{itemize}
\setlength{\itemsep}{0pt}
\setlength{\parskip}{0pt}
  \item Reparandum (1)
  \item Restart (1) 
\end{itemize}


\subsubsection{Adversarial}
\emph{Adversarial} instructions protect the model against basic manipulation and context-based elicitation of toxic or harmful behaviour. Additionally, they enhance the model's ability to comprehend more complex task formulations, such as embedded false theses or incomplete prompts. Subtypes of this category include:

\begin{itemize}
\setlength{\itemsep}{0pt}
\setlength{\parskip}{0pt}
  \item Ambiguous task (11)
  \item Confusing roleplay (9)
  \item Embedded false thesis (12)
  \item Identity implication (8)
  \item Instruction with no text (8)
  \item Link access request (11)
  \item Messy prompt (9)
  \item Negation prompt (10)
  \item Personal preference (12)
  \item Random text (14)
  \item Text with no instruction (9)
\end{itemize}

\subsubsection{Dialogue}\label{subsec:dialogue}
\emph{Dialogue} instructions serve multiple purposes. First of all, they target basic communication skills with natural examples of small talk and instruct the model in role-playing and adapting the response style to user demands. What is more, they integrate multiple task types within a single context, illustrate the adequate handling of context shifts, and incorporate previous conversation segments into new responses. The subtypes seem relatively unvaried, but most dialogues also incorporate multiple tasks described in other subsections:

\begin{itemize}
\setlength{\itemsep}{0pt}
\setlength{\parskip}{0pt}
  \item Multi-turn task (6)
  \item Roleplay (14)
  \item Small-talk (45)
  \item System prompt (15)
  \item Tasks combination (8)
\end{itemize}


\subsubsection{Formatting \& Visualization}
\emph{Visualization} instructions focus on the visual structure of generations. They include formatting guidelines and instructions for restructuring or presenting content in formats such as lists, tables, or graphs, i.e.:

\begin{itemize}
\setlength{\itemsep}{0pt}
\setlength{\parskip}{0pt}
  \item Content modification (24)
  \item Element addition (11)
  \item Graphs (14)
  \item Lists (18)
  \item Punctuation manipulation (6)
  \item Tables (16)
  \item Text casing modification (7)
\end{itemize}

\subsubsection{Data manipulation}
\emph{Data manipulation} instructions address the topic of data structures and fundamental data manipulation and analysis operations. This type enables the model to understand concepts such as output formats, conversion, basic modifications, or value extraction for common data formats, such as JSON or XML, cf. the subtypes:

\begin{itemize}
\setlength{\itemsep}{0pt}
\setlength{\parskip}{0pt}
  \item Conversion (12)
  \item Entity generation (14)
  \item Entity-based answer (27)
  \item Manipulation (18)
  \item Mistake correction (8)
  \item Return of value (15)
  \item Schemas (8)
  \item Structured output (38)
  \item Validation (13)
\end{itemize}

\subsubsection{Programming}
\emph{Programming} instructions acquaint the model with basic programming concepts, including foundational knowledge, code generation, and code comprehension. These instructions are designed to leverage pieces of information acquired during the pre-training phase. There are 5 subtypes belonging to this category:

\begin{itemize}
\setlength{\itemsep}{0pt}
\setlength{\parskip}{0pt}
  \item Code creation (21)
  \item Code explanation (24)
  \item Code rewriting (17)
  \item Concept explanation (15)
  \item Debugging (10)
\end{itemize}

\subsubsection{Chain of Thought}
\emph{Chain of Thought} instructions develop reasoning capabilities by focusing on answer explanations, step-by-step or how-to instructions, and generation with accompanying justifications. This category also targets a crucial LLM response ability based on one-shot and few-shot prompting techniques. The subtypes comprise:

\begin{itemize}
\setlength{\itemsep}{0pt}
\setlength{\parskip}{0pt}
  \item Few-shot (6)
  \item Generation with justification (11)
  \item How to (10)
  \item Multitask prompt (10)
  \item One-shot (8)
  \item Previous answer explanation (6)
\end{itemize}

\subsubsection{Translation}
\emph{Translation} instructions improve multilingual performance by working on language-focused tasks covered by the subtypes listed below, i.e. direct translation, translation with localization, proofreading, or explanation of concepts formulated in other languages:

\begin{itemize}
\setlength{\itemsep}{0pt}
\setlength{\parskip}{0pt}
  \item Context-based translation search (8)
  \item Instruction-Context language mismatch (8)
  \item Localization (11)
  \item Meaning explanation (7)
  \item Proof-reading (7)
  \item Translation (13)
  \item Translation matching (9)
\end{itemize}

\subsubsection{Identity}
\emph{Identity} instructions allow the model to establish a~sense of identity and affiliation. They encompass comprehensive information regarding its creation process, authorship, origin, purpose, and designation, as illustrated by the subtypes:

\begin{itemize}
\setlength{\itemsep}{0pt}
\setlength{\parskip}{0pt}
  \item Availability (11)
  \item Authorship (17)
  \item Competence (10)
  \item Creation (7)
  \item Name (9)
  \item Origin (7)
  \item Personality (8)
  \item Security (7)
\end{itemize}

\subsubsection{Thematic categorization}
\label{sec:appendix_thematic_categorization}
On top of the functional typology described in the previous sections, we also used a set of thematic areas to further categorize released samples according to topic:

\begin{itemize}
\setlength{\itemsep}{0pt}
\setlength{\parskip}{0pt}
  \item Art (14)
  \item Astronomy (5)
  \item Automotive (6)
  \item Biology (78)
  \item Chemistry (7)
  \item Computer science (163)
  \item Culinary (52)
  \item Culture (55)
  \item Ecology (4)
  \item Economy (19)
  \item Entertainment (85)
  \item Geography (59)
  \item History (48)
  \item Home (60)
  \item Hobby (4)
  \item Languages (185)
  \item Law and administration (31)
  \item Literature (50)
  \item Mathematics (15)
  \item Medicine (36)
  \item Other (73)
  \item Philosophy (5)
  \item Physics (8)
  \item Politics (42)
  \item Psychology (19)
  \item Religion (7)
  \item Society (169)
  \item Sports (26)
  \item Technology (87)
  \item Travel (25)
  \item Industry (20)
\end{itemize}

Each instruction is assigned a single main topic and up to two additional ones, to ensure proper descriptive quality.

\subsection{Synthetic instructions}

\subsubsection{Knowledge distilled}
\label{sec:appendix_synth_knowledge_distilled}
The objective of creating this instruction type was to represent a coherent set of best practices, such as proper formatting or style, in various contexts. This was intended to reinforce proper activations and prevent uneven performance in underrepresented domains. To achieve this, a taxonomy of high-level categories was established that was later systematically covered using similar meta-prompt guidelines:

\begin{itemize}
\setlength{\itemsep}{0pt}
\setlength{\parskip}{0pt}
  \item Artistic tasks
  \item Daily task management
  \item Data visualization
  \item Educational tasks
  \item Entertainment and media
  \item Expressing opinions and argumentation
  \item Medicine and health
  \item Problem-solving skills
  \item Project creation and management
  \item Socio-political contexts
  \item Technical tasks
\end{itemize}

\subsubsection{Context-injected}
\label{sec:appendix_synth_context_injected}

Open-source databases and annotated corpora were used to generate NLP-related instructions representing the following subtypes:

\begin{itemize}
\setlength{\itemsep}{0pt}
\setlength{\parskip}{0pt}
  \item Classification
  \item Extraction
  \item Keywords
  \item Knowledge alignment
  \item Lemmatization
  \item Morpho-syntactic tagging
  \item NER
  \item Semantic relations
  \item Sentiment analysis
  \item Summarization
  \item Text similarity
  \item Topic analysis
\end{itemize}

\subsection{Converted subsets}

{%
\onecolumn
\centering
\renewcommand*{\arraystretch}{1.15}
\small
\begin{longtable}{p{0.15cm}p{1.2cm}p{1.5cm}p{4cm}p{4cm}}
    
\toprule
\textbf{No} & \textbf{Source material} & \textbf{NLP task} & \textbf{Description} & \textbf{Example} \\
\midrule
1 & Curlicat \cite{varadi-etal-2022-introducing} & Key-word extraction & The dataset consists of abstracts of scientific papers and their respective multilingual keyword-sets. Keywords are being predicted based on the abstract text in different scenarios.   &  \textbf{Prompt:} \texttt{\{abstract text\}}\newline
Based on the text above generate a json file with a list of keywords in English.\\
\midrule
2 & DIABIZ \cite{diabiz_pezik_2} & Information extraction, text-classification & DIABIZ corpus is a dialogue corpus comprising recordings and annotated transcriptions of phone-based customer-agent interactions in several key business domains. Each interaction has a rich set of annotation items, including domain classification and intent annotation for selected turns of the dialogue. The latter describes any statement made by either the agent or the customer that has a defined purpose and prompts a defined response related to a specific business context. We transform the annotated dialogues into classification and extraction tasks in various scenarios.  &  \textbf{Prompt:} Identify the sentence in the presented conversation that matches the following description: \texttt{\{intent annotation\}} \newline Conversation:\texttt{\{conversation text\}}. \\
\midrule
3 & Paralela \cite{pezik_exploring_2016} & PL-EN, EN-PL translation & Paralela is a Polish-English parallel corpus covering a variety of manually and automatically aligned translations sourced from publicly available corpora. Based on the aligned segments we construct \texttt{pol-en} and \texttt{en-pol} translation tasks for text chunks of varying length.  &  \textbf{Prompt:} Translate into English the following text in Polish: \texttt{\{polish text\}} \\
\midrule
4 & Polish GEC datasets\footnote{\url{https://github.com/Ermlab/polish-gec-datasets}} & Error correction & Each dataset entry includes a sentence with errors and its corrected version. Content focus: common language errors, including syntax, orthography, and inflection errors in Polish. & \textbf{Prompt:} \texttt{Correct errors in the following sentence: \{sentence\}}\\
\midrule
5 & Polish book reviews dataset \cite{sentire2024} & Text classification and sentiment analysis & The dataset consists of material sourced from Polish literary and review blogs. Each entry includes a text classified as either a review or a non-review, along with sentiment annotations at both the sentence and whole-text levels. Sentiment annotations cover polarity (positive, negative, neutral) and intensity (weak, strong). The data has been processed into single-turn flat instructions and multi-turn dialogue instructions, where the model was prompted to classify the text, evaluate sentiment, and assess its intensity in various configurations.
 & \textbf{Prompt:} \texttt{You will receive a text from a blog. Your task is to assess whether the text qualifies as a review. Text: \{text\}} \textbf{Prompt:} \texttt{Identify the sentiment of the provided sentence. Choose from positive, negative, or neutral. Sentence: \{sentence\}}. \textbf{Prompt:} \texttt{Evaluate the intensity of the sentiment. Select either strongly positive or mildly positive.} \\
\midrule
6 & Lubimy czytać database & Question answering (QA) & Database of a community-based web service where users can rate, review, and discuss books. The data is converted into QA knowledge-driven prompts and answers, with multiple prompt variants. Note: No copyrighted material has been used in the subset & \textbf{Prompt:} \texttt{Who authored the book \{title\}?} \textbf{Prompt:} \texttt{What publishing house published the book \{title\}?}\\
\midrule
7 & Filmweb database & Question answering (QA) & Database of a community-based web service where users can rate, review, and discuss films. The data is converted into QA knowledge-driven prompts and answers, with multiple prompt variants, covering information on films, TV series, actors, and directors. Note: No copyrighted material has been used in the subset & \textbf{Prompt:} \texttt{When was \{actor\} born?} \textbf{Prompt:} \texttt{Name two films directed by \{director\}.}\\
\midrule
8 & Social media dataset \cite{banpl} & Anonymization & The task consists in anonymization of surnames and pseudonyms in linguistically challenging posts from social media. & \textbf{Prompt:} \texttt{In the text provided, anonymize only the surnames and nicknames, using the labels [surname] and [pseudonym] in place of the identified entities: \{text\}}\\
\midrule
9 & TLDR-PL abstractive summaries dataset & Text summarization and key words extraction & The TLDR-PL dataset features articles paired with human-annotated summaries and a list of 2-6 key words. Each summary is carefully crafted to represent 15\% of the original text, with a flexible deviation of ±10 words. The data has been processed into two-turn instructions, where the model is prompted to generate an abstractive summary and to extract 2 to 6 keywords that capture the essence of the text. & \textbf{Prompt:} \texttt{Summarize the following text. The abstract should contain 15\% of the initial text with a possible deviation of 10 words. \{text\}} \textbf{Prompt:} \texttt{I also need keywords, ranging from two to six.} \\
\midrule

10 & PolEval 2021: OCR correction dataset\footnote{\url{https://github.com/poleval/2021-ocr-correction}} \cite{kobylinski_etal_2021} & Error correction & The OCR correction dataset includes OCR-processed texts from Wikisources and their manually revised versions. The instructions are designed to fine-tune LLMs for proofreading, enabling them to correct OCR errors and typos and generate correct text outputs. & \textbf{Prompt:} Correct errors in the following scanned text: \texttt{\{text\}}.\\
\midrule
11 & Polish Summaries Corpus \cite{ogro:kop:14:lrec} & Text Classification and summarization & The PSC dataset consists of the articles from \textit{Rzeczpospolita} and three summaries of varying lengths for each article. Single- and multi-turn instructions are provided to guide LLMs to solve the summarization task. Additionally, single-turn instructions are also provided to solve a text classification task, in which the LLM is asked to predict whether a given text properly summarizes a passage. &  \textbf{Prompt:} Provide three different length summaries of this article \texttt{\{article\}} \textbf{Prompt:} Text: \texttt{\{text\}} Summary: \texttt{\{summary\}} Does the summary properly sums up the text? Answer concisely, yes or no. Correct answer:\\
\midrule
12 &  Corpus of Contemporary Polish \cite{kieras:etal:2024:kwjp} & Error correction & A small set of KWJP texts is intentionally altered with punctuation errors to create (incorrect-correct) text pairs. These pairs serve as the basis for instructions aimed at fine-tuning LLMs in correcting Polish punctuation. & \textbf{Prompt:} Check punctuation of this text: \texttt{\{text\}}\\
\midrule
13 & F19 \cite{kie:wol:lrec18} and Korba \cite{gru:etal:2022} & Text classification & The two corpora contain Polish texts from the 18th and 19th centuries. Each text is categorized into a historical period based on its writing date. The task is to fine-train LLMs to classify texts into the appropriate period using their linguistic characteristics.& \textbf{Prompt:} When was this text written? \texttt{\{text\}}\\
\midrule
14 & Polish Coreference Corpus \cite{10.1007/978-3-319-43808-5_17} & Coreference resolution & Prompting asks the model to return the text in a format with added coreference resolution markup, i.e., mentions spans and their corresponding entity numerical identifiers. & \textbf{Prompt:} Mark the coreference relations in the following text using square brackets and subscripts of the common reference - 
\texttt{[mention range]:index\_group} e.g. \texttt{[one of [Poles]:3]:2}.  Text: \texttt{\{text\}}\\
\midrule
15 & F19 \cite{kie:wol:lrec18} & Text modernization & The instructions are designed to fine-tune LLMs for modernising texts from the F19 corpus (19th-century Polish texts), into contemporary Polish. & \textbf{Prompt:} Adjust the text according to Polish spelling/orthographic rules. Text: \texttt{\{text\}}\\
\midrule
16 & Składnica \cite{wol:haj:21} & Error correction & The prompts provide a~sentence or short passage from Składnica constituency treebank (possibly containing an~automatically introduced syntactic error) and a~request for the model. Depending on the specific prompt, the model’s task is to either assess the grammaticality of the text or correct any errors. In part of the questions answer justifications are explicitly required. & \textbf{Prompt:} If the following text contains errors, correct them and justify: \texttt{\{text\}}. \\
\midrule
17 & SGJP \cite{sal:etal:15} & Common-sense knowledge extraction & The instructions concern examples of rare inflectional patterns in Polish extracted from the digital data of the SGJP grammatical dictionary. Each prompt gives a~word lemma and a~grammatical characteristic (case, number, etc.) and asks for inflected forms of the word matching the characteristic. The gold standard answers give all possible forms ($>1$ in case of lemma ambiguity). In case of ambiguity, where possible, the answer contains comments/explanations extracted from the dictionary data (glosses, stylistic qualifiers, named entity types). & \textbf{Prompt:} Provide all forms of \texttt{\{grammatical\_description\}} of the \texttt{\{part\_of\_speech\}} \texttt{\{lemma\}}. \\
\midrule
18 & Allegro Articles \cite{chrabrowa-etal-2022-evaluation} & Generation & Collection of articles from a popular Polish e-commerce marketplace -- allegro.com. They are mostly product reviews and shopping guides. The task is to write an article for a given title. & \textbf{Prompt:} Write an article of about \texttt{\{length\}} words on the given title: \texttt{\{title\}} \\
\midrule
19 & PolQA \cite{rybak-etal-2024-polqa} & Question answering (QA) & Collection of trivia questions and short answers collected from TV shows, online quizzes, etc. Each question is linked to a Wikipedia article that contains the correct answer. The dataset is used for three tasks: closed-book QA, open-book QA, and reranking. & \textbf{Prompt:} Decide whether the passage answers the question. Question: \texttt{\{question\}} Passage: \texttt{\{passage\}}\\
\midrule
20 & PoQuAD \cite{10.1145/3587259.3627548} & Question answering (QA) & A SQuAD-like dataset for Polish QA. It consists of Wikipedia articles and manually written questions. The dataset is used for two tasks: open-book QA and reranking. & \textbf{Prompt:} Write a short answer based on a given passage. Question: \texttt{\{question\}} Passage: \texttt{\{passage\}} \\
\midrule
21 & DYK \cite{marcinczuk2013open} & Question answering (QA) & The Did You Know (pol. \textit{Czy wiesz?}) dataset consists of human-annotated question-answer pairs. The task is to predict if the answer is correct. Examples were processed into instructions with several prompt variants. & \textbf{Prompt:} Question: \texttt{\{question\}} Suggested Answer: \texttt{\{answer\}} Is the suggested answer correct? Respond concisely with either True or False. Answer: \\
\midrule
22 & PolEmo2 \cite{kocon-etal-2019-multi} & Text classification and sentiment analysis & A human-annotated dataset of online Polish reviews from hotels, medicine, university and products domains. The task is to predict the sentiment contained in the given text. The data were converted into instructions with several prompt variations. & \textbf{Prompt:} Opinion: \texttt{\{text\}} What type of sentiment does the given opinion express? Negative, neutral, ambivalent or positive? \\
\midrule
23 & Polish Paraphrase Corpus \cite{9945218} & Text classification and paraphrasing & A classification dataset for paraphrase identification. It contains manually labelled sentence pairs drawn from Wikipedia, Polish news articles, Taboeba, and Polish version of SICK dataset. The dataset's author manually altered some sentences to balance the classes. We processed the data into instructions with several prompt variations. & \textbf{Prompt:} Question: What is the relationship between the given sentences? Sentence 1: \texttt{\{sentence1\}} Sentence 2: \texttt{\{sentence2\}} Possible Answers: A. They have a similar meaning. B. They have different meanings. C. They mean exactly the same thing. Correct Answer: \\
\midrule
24 & CDSC-E \cite{wroblewska2017polish} & Text classification and textual entailment recognition & It contains Polish sentence pairs,  human-annotated for semantic entailment. The task is to predict if the relation between the sentence pairs is neutral, entailment, or contradiction. We converted the data into instructions with several prompt formats. & \textbf{Prompt:} Sentence A: \texttt{\{sentenceA\}} Sentence B: \texttt{\{sentenceB\}} Instruction: Determine the relationship between the given pair of sentences. Possible Answers: entailment, contradiction, neutral. Answer concisely without elaboration. Answer: \\
\midrule
25 & 8tags \cite{dadas-etal-2020-evaluation} & Text classification & It contains Polish social media headlines classified into topics. The headlines were collected from the Polish platform \textit{wykop.pl}, where users can assign category tags to posts. The task is to classify a given text into one of eight possible topics. The data were converted into instructions, utilising different prompt formats. & \textbf{Prompt:} Title: \texttt{\{title\}} Which category best fits the given title? Film, History, Food, Medicine, Automotive, Work, Sport, or Technology? \\
\midrule
26 & NKJP-NER \cite{przepiorkowski2012narodowy} & Text classification and Named Entity Recognition (NER) & The dataset contains extracted sentences from the National Corpus of Polish (pol. \textit{Narodowy Korpus Języka Polskiego} -- NKJP). Each text may contain only one type of named entities. The task is to predict the named entity type, if any. We processed the data into instructions with several prompt variations. & \textbf{Prompt:} Sentence: \texttt{\{sentence\}} Instruction: Select the type of named entity from the options below if a named entity appears in the sentence above. Respond with only A, B, C, D, E or F. Possible Answers: A - Person Name B - Time C - Organization Name D - No Entity E - Geographical Name F - Place Name Correct Answer: \\
\midrule
27 & KPWr \cite{broda-etal-2012-kpwr} & Named Entity Recognition (NER) & The dataset contains extracted sentences from the Polish Corpus of Wrocław University of Technology (pol. \textit{Korpus Języka Polskiego Politechniki Wrocławskiej} -- KPWr) manually annotated with named entities. We processed the data into instructions with several prompt variations. & \textbf{Prompt:} Provide the identifying units present in the text \texttt{\{text\}}. \textbf{Prompt:} Given the provided list of proper name types \texttt{\{types\_list\}}, provide their examples from the text \texttt{\{text\}}. \textbf{Prompt:} What identifying units can be found in the text \texttt{\{text\}}? \textbf{Prompt:} Given the text fragment  \texttt{\{text\}}, extract all types of proper names present along with the words/phrases representing them. The possible types are \texttt{\{types\_list\}} \\
\midrule
28 & Schema Guided Dialogue State Tracking \cite{rastogi2020towards} & Task-oriented conversation completion  & The instructions are based on task-oriented conversations from the SG-DST dataset, where the objective is to complete the assistant's turns in a dialogue based on the given task-specific dialogue context, which includes domains such as banking, events, media, calendars, travel, and weather. & \textbf{Prompt:} Based on the previous fragment of the dialogue between the user and the system: \texttt{\{dialogue\}} propose the next part of the dialogue that aligns with the following external data: \texttt{\{service\_results\}},
    \textbf{Prompt:} Based on the fragment of the dialogue between the user and the system and the data obtained from the API, propose the continuation of the conversation. Dialogue: \texttt{\{dialogue\}} The API data: \texttt{\{service\_results\}},
    \textbf{Prompt:} Continue the dialogue based on the previous conversation between the user and the system as well as the following external information: \texttt{\{dialogue\}} External information: \texttt{\{service\_results\}},
    \textbf{Prompt:} Continue the conversation, taking into account the previous dialogue between the user and the system as well as the following external data: \texttt{\{dialogue\}} External data: \texttt{\{service\_results\}},
    \textbf{Prompt:} Based on the past conversation between the user and the system as well as the data from the API, create the next part of the conversation: \texttt{\{dialogue\}} The API data: \texttt{\{service\_results\}} \\
    \midrule
29 & Schema Guided Dialogue State Tracking \cite{rastogi2020towards} & Attribute value extraction (dialogue state tracking) & The collection contains labelled dialogues. Each turn of dialogue is annotated with the attributes and values of the user's utterances, which are later used in the search. We processed the data into instructions with several prompt variations. & \textbf{Prompt:} In the given text \texttt{\{text\}} find information on the specified topic: \texttt{\{attribute\}}. If this information is not present, return 'null',
\textbf{Prompt:} Based on the passage: \texttt{\{text\}} provide: \texttt{\{attribute\}}. If such information is not available, return 'null',
\textbf{Prompt:} Given the text \texttt{\{text\}} extract information about: \texttt{\{attribute\}},
\textbf{Prompt:} Find information about \texttt{\{attribute\}} in the following text: \texttt{\{text\}} If this information is missing, return 'null',
\textbf{Prompt:} Does the given text \texttt{\{text\}} contain information about: \texttt{\{attribute\}} If so, provide it. \\
\midrule
30 & Unified Sense Inventory for Word Sense Disambiguation in Polish \cite{10.1007/978-3-031-08754-7_70} & Word Sense Disambiguation & The instructions are based on the comprehensive evaluation benchmark for Polish Word Sense Disambiguation task. The benchmark consists of 7 distinct datasets with sense annotations based on plWordNet--4.2. We processed the data into instructions with several prompt variations. & \textbf{Prompt:} Given the sentence: \texttt{\{text\}}, how would you define the following word: \texttt{\{word\}}?,
\textbf{Prompt:} Provide the definition of the highlighted word in this text \texttt{\{context\}},
\textbf{Prompt:} Provide the definition of the word: \texttt{\{word\}} based on the following context of its usage: \texttt{\{context\}},
\textbf{Prompt:} Based on the sentence \texttt{\{text\}}, how would you describe the meaning of the following word: \texttt{\{word\}}?
\textbf{Prompt:} Does the word \texttt{\{word} in the text: \texttt{\{text\}} has the same meaning as in this one: \texttt{\{text2\}},
\textbf{Prompt:} Does the provided definition \texttt{\{definition\}} describe the word: \texttt{\{word\}} in the context of \texttt{\{text\}}?
\textbf{Prompt:} Provide the definition of the word \texttt{\{word\}},
\textbf{Prompt:} What are possible definition of the word \texttt{\{word\}}?,
\textbf{Prompt:} Provide the most common definition of the word: \texttt{\{word\}}. \\
\midrule
31 & VeSNet (EuroVoc, GEMET, Wikidata) \cite{janz2021mapping} & Word Sense Disambiguation & The instructions are based on terminology definitions from a network of lexical resources resulting from the merge of Polish-English WordNet (PEWN) with several existing large electronic thesauri from the Linked Open Data cloud (EuroVoc, GEMET, Wikidata). We processed the data into instructions with several prompt variations. & \textbf{Prompt:} Please provide me with the definition of the term \texttt{\{word\}}?,
\textbf{Prompt:} What is the meaning of the expression \texttt{\{word\}}?,
\textbf{Prompt:} What does the expression \texttt{\{word\}} mean?,
\textbf{Prompt:} What is \texttt{\{word\}}?,
\textbf{Prompt:} Please provide me with the meaning of the following expression \texttt{\{word\}},
\textbf{Prompt:} How would you define the term \texttt{\{word\}}? \\
\midrule
32 & CST Directed (Podcast, WNLI, SNLI-REF, WUT-REF) \cite{10.1007/978-3-031-70816-9_21} & Relationship Extraction & The instructions are based on a collection of corpora manually annotated with the relations between sentences (CST, NLI). We processed the data into instructions with several prompt variations. & \textbf{Prompt:} Possible types of relations between sentences are: \texttt{\{relation\_list\}}. What relationship exists between the following sentences \texttt{\{s1\}} and \texttt{\{s2\}}?,
\textbf{Prompt:} Given the dictionary of relationship types, where the key is the name of the relationship type and the value is its definition: \texttt{\{relation\_dictionary\}} determine what relationship exists between the given sentences: a) \texttt{\{s1\}} b) \texttt{\{s2\}},
\textbf{Prompt:} For the two sentences, \texttt{\{s1\}} and \texttt{\{s2\}} provide the type of semantic relationship between them (if one exists). The type of relationship should be chosen from the list: \texttt{\{relation\_list\}},
\textbf{Prompt:} Provide the type of semantic relationship between the sentences \texttt{\{s1\}} and \texttt{\{s2\}}. Possible relationship types along with their definitions are: \texttt{\{relation\_bullet\_list\}},
\textbf{Prompt:} Among the possible relationship between sentences are: \texttt{\{relation\_bullet\_list\}}. What kind of connection exists between sentence \texttt{\{s1\}}, and sentence \texttt{\{s2\}}? \\
\midrule
33 & Polish CBD \cite{ptaszynski2023expert} & Text classification and hate speech detection & An expert-annotated dataset containing annotations of cyberbullying and hate-speech of Polish texts. The task is to predict whether the given text belongs to one of the hate speech categories. The data were converted into single-turn flat instructions with several prompt variations. & \textbf{Prompt:} Statement: \texttt{\{text\}} Which of the following categories best describes the given statement? Harmless, mockery, insult, insinuation, threat, harassment. Respond concisely with a single word. Category: \\
 \\
\bottomrule
\caption{Examples of datasets converted to instruction.}
\label{tab:converted_taks_examples}
\end{longtable}
}






\end{document}